\documentclass[letterpaper]{article} 
\usepackage[submission]{aaai25}  
\usepackage{times}  
\usepackage{helvet}  
\usepackage{courier}  
\usepackage[hyphens]{url}  
\usepackage{graphicx} 
\usepackage{graphicx} 
\usepackage{natbib}  
\usepackage{caption} 
\usepackage[dvipsnames]{xcolor}
\usepackage{algorithm}
\usepackage{algpseudocode}

\usepackage{newfloat}
\usepackage{listings}
\DeclareCaptionStyle{ruled}{labelfont=normalfont,labelsep=colon,strut=off} 
\floatstyle{ruled}
\newfloat{listing}{tb}{lst}{}
\floatname{listing}{Listing}

\usepackage{amsmath} 
\usepackage{amssymb}
\usepackage{booktabs} 
\usepackage{multirow} 
\usepackage{colortbl} 
\usepackage{arydshln} 
\usepackage{xcolor} 
\usepackage{pdflscape} 
\usepackage{subcaption} 
\usepackage{soul} 
\usepackage{enumitem} 
\usepackage{siunitx} 
\usepackage{microtype} 
\usepackage{diagbox}
\usepackage{colortbl}
\usepackage{xcolor}
\usepackage{pgfplots}
\usepackage{todonotes}
\usepackage{diagbox} 
\pgfplotsset{compat=1.17}

\usepackage{hyperref}

\definecolor{ForestGreen}{RGB}{34,139,34}

\begin{document}

\title{\textit{Multilingual Needle in a Haystack:} Investigating Long-Context Behavior of Multilingual Large Language Models}
\author {
    Amey Hengle\textsuperscript{\rm 1},
    Prasoon Bajpai\textsuperscript{\rm 1},
    Soham Dan\textsuperscript{\rm 2},
    Tanmoy Chakraborty\textsuperscript{\rm 1} 
}
\affiliations {
    \textsuperscript{\rm 1}Indian Institute of Technology Delhi, India;  
    \textsuperscript{\rm 2}Independent\\
    \{ameyhengle22, prasoonbajpai786, sdan021\}@gmail.com, tanchak@iitd.ac.in
}

\maketitle

\begin{abstract}
While recent large language models (LLMs) demonstrate remarkable abilities in responding to queries in diverse languages, their ability to handle long multilingual contexts is unexplored. As such, a systematic evaluation of the long-context capabilities of LLMs in multilingual settings is crucial specifically in the context of information retrieval. To address this gap, we introduce the \textit{MultiLingual Needle-in-a-Haystack} (MLNeedle) test, designed to assess a model's ability to retrieve relevant information (the \textit{needle}) from a collection of multilingual distractor texts (the \textit{haystack}). This test serves as an extension of the multilingual question-answering task, encompassing both monolingual and cross-lingual retrieval. We evaluate four state-of-the-art LLMs on MLNeedle. Our findings reveal that model performance can vary significantly with language and needle position. Specifically, we observe that model performance is the lowest when the needle is (i) in a language outside the English language family, and (ii) located in the middle of the input context. Furthermore, although some models claim a context size of $8k$ tokens or greater, none demonstrate satisfactory cross-lingual retrieval performance as the context length increases. Our analysis provides key insights into the long-context behavior of LLMs in multilingual settings to guide future evaluation protocols. To our knowledge, this is the first study to investigate the multilingual long-context behavior of LLMs.
\end{abstract}

\section{Introduction}

In recent years, Large Language Models (LLMs) have demonstrated remarkable capabilities across a wide range of natural language processing tasks, including text generation, translation, and question-answering. A critical aspect of these models is their ability to handle long input contexts effectively — a capability that is essential for applications such as document summarization, long-form content generation, and multi-turn dialogue systems \cite{petroni-contextaffectslanguagemodels-2020, Lee_2022, thoppilan-lamdalanguagemodelsdialog-2022}. This ability directly impacts the relevance and accuracy of LLM responses over long inputs \cite{khandelwal-etal-2018-sharp, mallen2023trustlanguagemodelsinvestigating, shaham2023zeroscrollszeroshotbenchmarklong, kandpal2023largelanguagemodelsstruggle}.

A recent study  \cite{hsieh2024-Ruler} has shed light on the potential and limitations of LLMs in handling extended sequences. They evaluate various attention mechanisms and model architectures, revealing that while some models can technically manage long contexts, their effective utilization of this capacity is often suboptimal. Furthermore, \citet{Liu2023-LostITM} highlighted a major problem faced by LLMs while handling long contexts: a marked decline in performance when the relevant information is located in the middle of a long input context. The performance curve is characteristically ``U-shaped", where models exhibit better accuracy when the relevant information is at the beginning or end of the context, underscoring the challenges LLMs face in maintaining attention and relevance throughout an entire input sequence. This phenomenon, dubbed the \textit{lost-in-the-middle} problem, suggests that current LLMs are not yet fully equipped to handle long contexts robustly and reliably, particularly in tasks that require retrieval of dispersed information \cite{Liu2023-LostITM, ivgi-etal-2023-efficient, wang-Multimodal-NIAH-2024}.

\begin{figure}[t]
\centering \includegraphics[width=0.95\columnwidth]{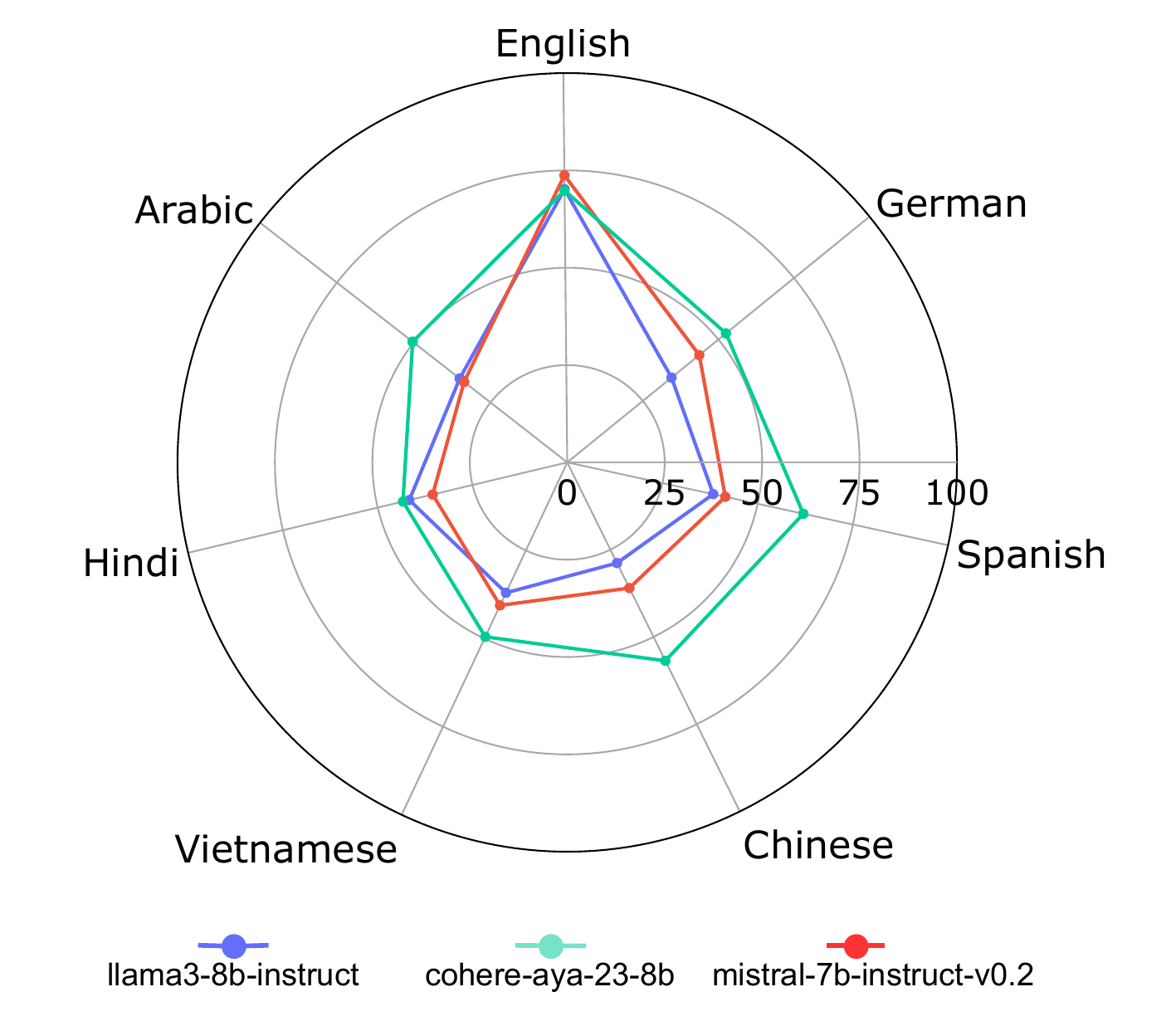}
\caption{Monolingual long-context performance (accuracy in radial axis) for various LLMs averaged across different context sizes (4K, 8K, 16K, and 32K). We observe a considerable drop in performance for all languages except English, suggesting that multilingual LLMs struggle to process non-English (or non-Latin) long input contexts.}
\label{fig:figure0}
\vspace{-5mm}
\end{figure}

While these papers shed some light on the long-context capabilities of LLMs, a significant research gap remains: most existing benchmarks and evaluations have focused exclusively on monolingual English settings. This leaves a critical question unanswered: How do LLMs perform when the long input contexts are multilingual, particularly when the context is in a low-resource, non-Latin language? As shown in Figure \ref{fig:figure0}, LLMs usually perform poorly while handling non-English long contexts. Furthermore,  multilingual and cross-lingual contexts introduce additional complexities, such as varied syntax, grammar, and semantic nuances, which can significantly affect a model's retrieval and processing capabilities. 
In this paper, we address this gap by analyzing how LLMs process and retrieve information from long multilingual contexts. Specifically, we introduce the \textbf{MultiLingual Needle-in-a-Haystack (MLNeedle)} test, which extends the multilingual question-answering task to assess the LLM's ability to locate and extract relevant information (the \textit{needle}) from a large collection of multilingual distractor texts (the \textit{haystack}). Our experiments systematically vary the language and position of the needle within the haystack to evaluate the robustness of several state-of-the-art LLMs in handling multilingual long contexts.  we make the following contributions\footnote{The source code and datasets are available at \url{https://github.com/AmeyHengle/multilingual-needle-in-a-haystack}}: 

\begin{itemize}[]
\item We introduce the Multilingual Needle in a Haystack (MLNeedle) test, a first step towards systematically evaluating the long-context capabilities of multilingual LLMs. MLNeedle assesses model performance across seven languages in both monolingual and cross-lingual settings, providing a comprehensive benchmark for future research.
\item We conduct a series of controlled experiments to examine how changes in the language and position of the needle affect model performance. Our findings reveal that LLM performance is highly sensitive to both the \textit{language} and \textit{position} of the needle in the haystack. 
\item We demonstrate the relative robustness of LLMs to variations in the language of distractor passages, indicating that the key challenges lie in how LLMs process and retrieve the \textit{needle} from diverse linguistic environments.
\item We perform several ablation studies to understand the role of temperature sampling, instruction tuning and the choice of evaluation metric on performance.
\end{itemize}

\begin{figure*}[t]
\includegraphics[width=\textwidth,keepaspectratio,trim={0 0 0 0},clip]{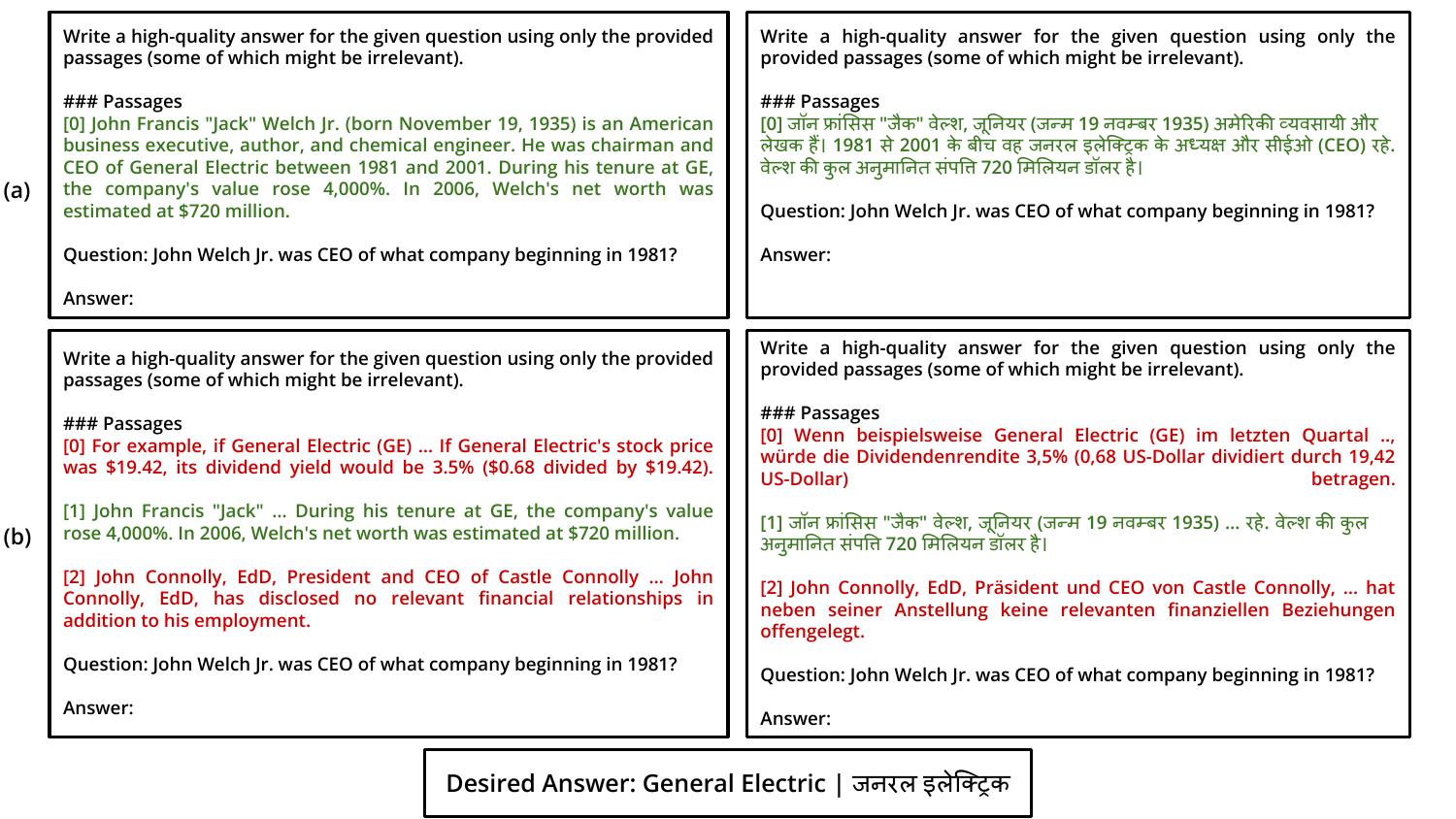}
\caption{Example of a multilingual question-answering input from the MLNeedle dataset. (Top) There are no distractor documents and the same needle (highlighted in \textcolor{OliveGreen}{green}) is present in English (Left) and Hindi (Right);  (Bottom) There are distractor documents present (highlighted in \textcolor{red}{red}) and the same needle is present in English (Left) and Hindi (Right).}
\label{fig:figure5}
\vspace{-5mm}
\end{figure*}

\begin{table*}[t]
\begin{center}
\begin{tabular}{l|cccccccccc}
\toprule
\textbf{Model} & \textbf{Claimed Length} & \textbf{Effective Length} & \textbf{Baseline} & \textbf{4K} & \textbf{8K} & \textbf{16K} & \textbf{32K} & \textbf{Avg.} \\ 
\midrule
Llama2-7B-Chat & 4K & {$<$4K} & \underline{$0.335$} & {$0.171$} & $-$ & $-$ & $-$ & $0.253$  \\
Llama3-8B-Instruct & 8K & {4K} & \underline{$0.622$} & \underline{$0.479$} & $0.295$ & $-$ & $-$ & $0.465$ \\
Cohere-Aya-23-8B & 8K & {4K} & \underline{$0.700$} & $0.460$ & $0.449$ & $-$ & $-$ & $0.536$ \\
Mistral-7B-Instruct-v0.2 & 32K & {8K} & \underline{$0.579$} & \underline{$0.485$} & \underline{$0.455$} & {$0.427$} & {$0.397$} & $0.469$ \\
\bottomrule
\end{tabular}%
\end{center}\caption{Long-context performance of selected models on the MLNeedle test. Models are evaluated for context lengths ranging from 4K to 32K. Each score is determined by averaging the accuracy of MLNeedle's multilingual question-answering task. The effective length is the maximum context length beyond which the model's performance decreases by more than $25\%$ from its baseline performance. The accuracy values within $25\%$ of baseline performance are \underline{underlined}, showcasing the effective length.  For all models, we observe that the claimed context size differs from the effective context size. }
\label{tab:main_results}
\end{table*}

\section{MultiLingual Needle in a Haystack}
\label{sec:MLNeedle_setup}
In this section, we introduce our MultiLingual Needle in a Haystack (MLNeedle) benchmark. As our goal is to better understand how LLMs process multilingual input contexts, we analyze the performance of the model on a multilingual question answering task, which requires a model to find relevant information (the \textit{needle}) from the input context (the \textit{haystack}) to answer the given question. Specifically, we conduct experiments where we systematically change (i) the position of the needle, (ii) the language of the needle, and (iii) the language of the haystack, and study the effect on performance. If LLMs are able to use information from long multilingual contexts, their performance should remain relatively stable regardless of changes in language or needle position.

\subsection{Experimental Setup}
\label{sec:experimental_setup}
In the multilingual question-answering task, the model is provided with a question $Q$ to answer and $K$ documents. Among these documents, exactly one contains the correct answer to the question $Q$, while the remaining $K-1$ \textit{distractor} documents do not. We denote the document containing the correct answer as $N$ (the \textit{needle}) and the $K-1$ distractor documents as $H$ (the \textit{haystack}). Figure \ref{fig:figure5} offers an overview of our evaluation setup using a randomly sampled example from the MLNeedle dataset (details in Section \ref{sec:MLNeedle_Dataset}). As illustrated in the figure, we systematically change the language of both the needle (highlighted in green) and the distractor documents (highlighted in red) during our experiments. The content of $N$ and $H$ remains unchanged, only the language varies. Therefore, the LLM's performance should ideally not fluctuate due to these changes. We also vary the position of $N$ within the input context, as highlighted in Figure \ref{fig:figure6} (Appendix), to understand the effect on the LLM's ability to retrieve. The language of the question is kept fixed in English.

\subsection{The MLNeedle Dataset}
\label{sec:MLNeedle_Dataset}
We instantiate MLNeedle with the MLQA dataset \cite{lewis2020-MLQA}, which consists of over 5K extractive question-answer instances across seven languages (English, Arabic, German, Spanish, Hindi, Vietnamese, and Simplified Chinese) in the SQuAD \cite{rajpurkar-etal-2016-SQUAD} format. We choose MLQA because of its aligned dataset structure, where each question-answer pair is present in multiple languages. Specifically, for each question-answer instance, there are corresponding versions in at least four different languages, allowing for a direct comparison of how the same question is answered across various linguistic contexts. We highlight this in Figure \ref{fig:figure5} (a), where for the given question, the needle document with the correct answer can be presented in both English as well as Hindi. This setting allows us to systematically study the effect of changing the language of the needle $N$.

\paragraph{Constructing the Haystack ($H$).} In MLNeedle, we collect the $K-1$ distractor documents for each question-answer pair using the following procedure: we use Wikipedia passages from mMARCO \cite{bonifacio2022-mMARCO}, a well-known multilingual passage ranking dataset, as the source of the distractor documents. For each question-answer pair, we use multilingual sentence-BERT \cite{reimers-2020-multilingual-sentence-bert} to retrieve $K-1$ documents from mMARCO that are most relevant to the question but do not contain the answer. Appendix \ref{sec:noise_retrieval_process} provides a detailed explanation of our retrieval system. In the final input context, we arrange these distractor documents in order of decreasing relevance. As highlighted in Table \ref{tab:main_results}, we conduct experiments across different context sizes ranging from 4K up to 32K tokens. To modulate the context length, we simply increase or decrease the number of distractor documents in $H$. Further details on varying the context length can be found in Appendix \ref{sec:controlling_context_length}.

\paragraph{Positioning the Needle ($N$).} We modulate the position of the relevant information within the input context by placing the document with the correct answer ($N$) at either the \textit{start}, \textit{middle}, or \textit{end} of $H$ (Figure \ref{fig:figure6} in Appendix), following prior experimental setups \cite{Liu2023-LostITM, ivgi-etal-2023-efficient}.

\subsection{Models}
We analyze several state-of-the-art open-source language models for our evaluation, spanning $32K$, $8K$ and $4K$ models.  We report \textbf{Mistral-7B-Instruct-v0.2} \cite{Jiang-Mistral-2023}, which features a maximum context length of $32,768$ tokens and is a multilingual model capable of understanding and generating text in multiple languages. It makes use of a unique positional encoding method, ALiBi \cite{press-Alibi-2022}, to effectively manage long-range dependencies. Next, we evaluate \textbf{Cohere-Aya-23-8B} \cite{AYA-23-openweight-2024}, which also supports multilingual capabilities and has a context size of $8,192$ tokens. This model is designed to perform well across various language tasks. We include \textbf{Llama3-8B-Instruct} \cite{llama3modelcard}, an instruction fine-tuned version of the Llama3 base model \cite{dubey2024-llama3}, which supports a context size of $8,192$ tokens. This model is optimized for following instructions and engaging in open-ended dialogue. Lastly, we evaluate\textbf{Llama2-7B-Chat} \cite{touvron2023-Llama2}, which has a maximum context length of $4,096$ tokens. 

\subsection{Evaluation Metric}
\label{sec: evaluation_metric}
As we evaluate the model's ability for retrieval in question-answering, we use \textbf{exact accuracy} \cite{kandpal2023-LLMs-longtail-knowledge, mallen2023-When-not-trust-LLMs} as our primary evaluation metric. Additionally, in our ablation study (see Section \ref{sec:ablation_analysis}), we report \textbf{existence accuracy} \cite{wang-Multimodal-NIAH-2024}, which evaluates the secondary task of determining whether relevant information is present within the input context. Below, we formally define each of these metrics:

\begin{itemize}[noitemsep,topsep=0pt]
\item \textbf{Exact Accuracy} measures the proportion of samples where the model's predicted output contains any of the correct answers, as specified in the MLQA dataset. This metric checks whether the ground-truth answer is contained in the model's predictions.
\item \textbf{Existence Accuracy} is the proportion of samples where the model correctly identifies whether a document containing the correct answer exists within the input context.

\end{itemize}

We define the evaluation prompt templates for respective tasks in Appendix \ref{sec:prompt_templates}. For consistency, we use the same evaluation prompts across all models in our experiments. Furthermore, for each input prompt, the model generates a prediction, which may be in a different language from the ground truth. To ensure accurate comparison, we translate the prediction to English (since each instance has a golden answer in English in MLQA) using Google Translate API\footnote{\href{https://cloud.google.com/translate/docs}{Google Translate API}}. We then use the translated predictions and the ground truth to calculate Exact Accuracy as defined earlier. Further details of our evaluation framework are given in Appendix \ref{appendix:evaluation_framework}. 

\begin{figure*}[t]
\includegraphics[width=\textwidth,keepaspectratio,trim={0 0 0 0},clip]{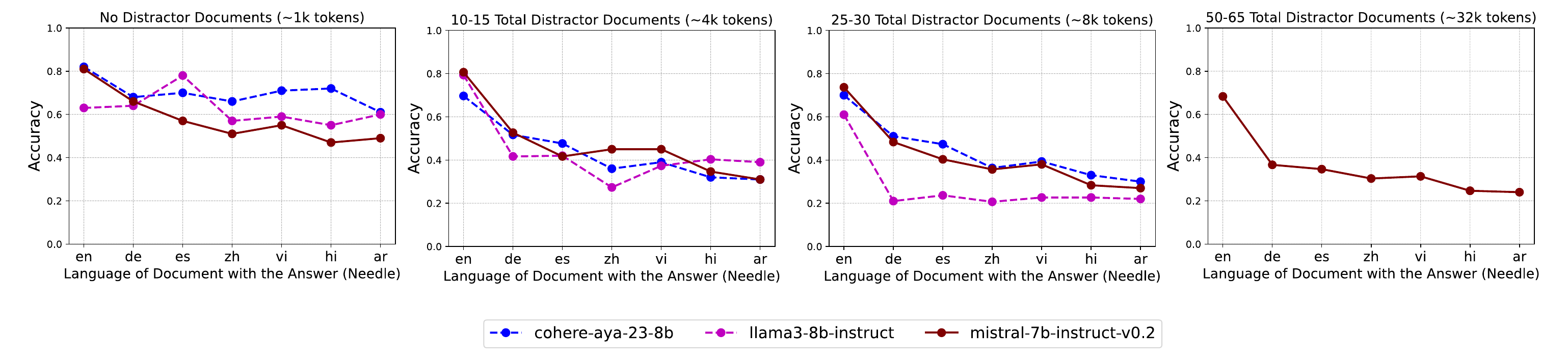}
\caption{Effect of changing the language of answer document (needle).}
\label{fig:figure3}
\vspace{-5mm}
\end{figure*}

\section{Experimental Results}
We run experiments with context sizes ranging from 4K to 32K tokens to compare the performance of various models on MLNeedle \footnote{All generation experiments were conducted using fixed sampling parameters to control the randomness and diversity of the generated responses. Specifically, we set the temperature to $0.7$ and used top-$k$ sampling with $k=50$. These parameters were kept constant across all selected models to ensure a fair comparison of their performance on the MLNeedle test.}. Table \ref{tab:main_results} summarises each model's performance, averaged across the seven languages in MLNeedle. As the context size increases, all models show a significant drop in performance. To evaluate the maximum context size that each model can handle effectively, we define the \textit{effective length} as the maximum context length at which the model's performance does not decrease by more than $25\%$ from its baseline accuracy \footnote{Baseline accuracy refers to the model's performance when there are no distractors in the input context (see Figure \ref{fig:figure5}(a)).}. Our findings show that all models struggle with longer contexts, exhibiting significant drops in accuracy beyond their effective lengths. Figure \ref{fig:figure0} shows the monolingual (both needle and distractors in the same language) performance of different models on the MLNeedle test. Models consistently perform better in English than non-English languages. In the following sections, we will investigate the effect of modulating the position and language of the needle.

\subsection{Effect of Changing the Needle Position}
\label{sec:effect_of_changing_needle_position}
In this experiment, we evaluate how the position of the needle ($N$) within the input context affects retrieval performance. We systematically place the needle at the {\em start}, {\em middle}, and {\em end} of the input context to understand how its position influences the model's ability to accurately retrieve the correct information. As shown in Figure \ref{fig:figure4}, the model performs best when the needle is placed at the beginning or end of the input context. These findings extend \citet{Liu2023-LostITM} to multilingual settings. The preference for starting and ending positions to retrieve relevant information indicates a potential weakness in the model's ability to maintain effective attention throughout the input sequence \cite{hsieh-found-in-the-middle-2024}, and exploring this in the multilingual setting is an important future work. 

\subsection{Effect of Changing the Needle Language}
In this experiment, we investigate how changing the language of the document containing the correct answer ($N$) affects the retrieval performance of selected models. To isolate the effects of needle language, we keep the distractor passages ($H$) in English.  The results, illustrated in Figure \ref{fig:figure3}, show that LLMs perform best when $N$ is either in English or in a language that is \textit{close} to English. However, as we move away from the Latin languages family, we notice a significant drop in performance. The decrease in performance is most noticeable when $N$ is presented in languages significantly different from English, such as Chinese and Arabic. When the language of $N$ is changed from English to German or Spanish, both of which are \textit{close} to English, the performance drop is moderate. This suggests that the model is relatively effective in processing content in languages that share similarities with English. On the other hand, the drop in performance is more pronounced when we change the language of $N$ from English to non-Latin languages such as Chinese and Arabic. The substantial drop in performance indicates that the models struggle to effectively process and retrieve the same content when presented in these linguistically distant languages.  

Our findings suggest that although the content of $N$ remains unchanged, LLMs display considerable variability in their ability to retrieve the correct information depending on the language of $N$. This inconsistency underscores a critical limitation of current LLMs: \textit{The retrieval capability is heavily influenced by the language in which the content is presented}. The models performs better for high-resource languages like English, but their performance diminishes in lower-resource languages such as Hindi.

\subsection{Effect of Changing the Haystack Language}
Here, we investigate how varying the language of the distractor documents, $H$, impacts the retrieval performance of LLMs. We keep the language of the needle constant, as English, and then systematically change the language of the haystack.  Table \ref{tab:pairwise_accuracy} shows that changing the haystack language from English to Arabic does not significantly affect the model's performance. This observation suggests that LLMs are relatively robust to changes in the language of non-relevant information, indicating that the model can focus on and retrieve the relevant content without being easily confused by the language of distractor passages. In contrast to the significant decline in performance observed when changing the language of the needle, the models' performance appears comparatively more stable when only the language of the distractor documents is altered. This suggests that the retrieval task's difficulty is more sensitive to the language of the needle than to the language of the haystack. LLMs focus on the content of the needle, rather than being distracted by the language of the haystack, highlighting its ability to prioritize relevant information effectively in this cross-lingual setting.

\begin{figure*}[t]
\includegraphics[width=\textwidth,keepaspectratio,trim={0 0 0 0},clip]{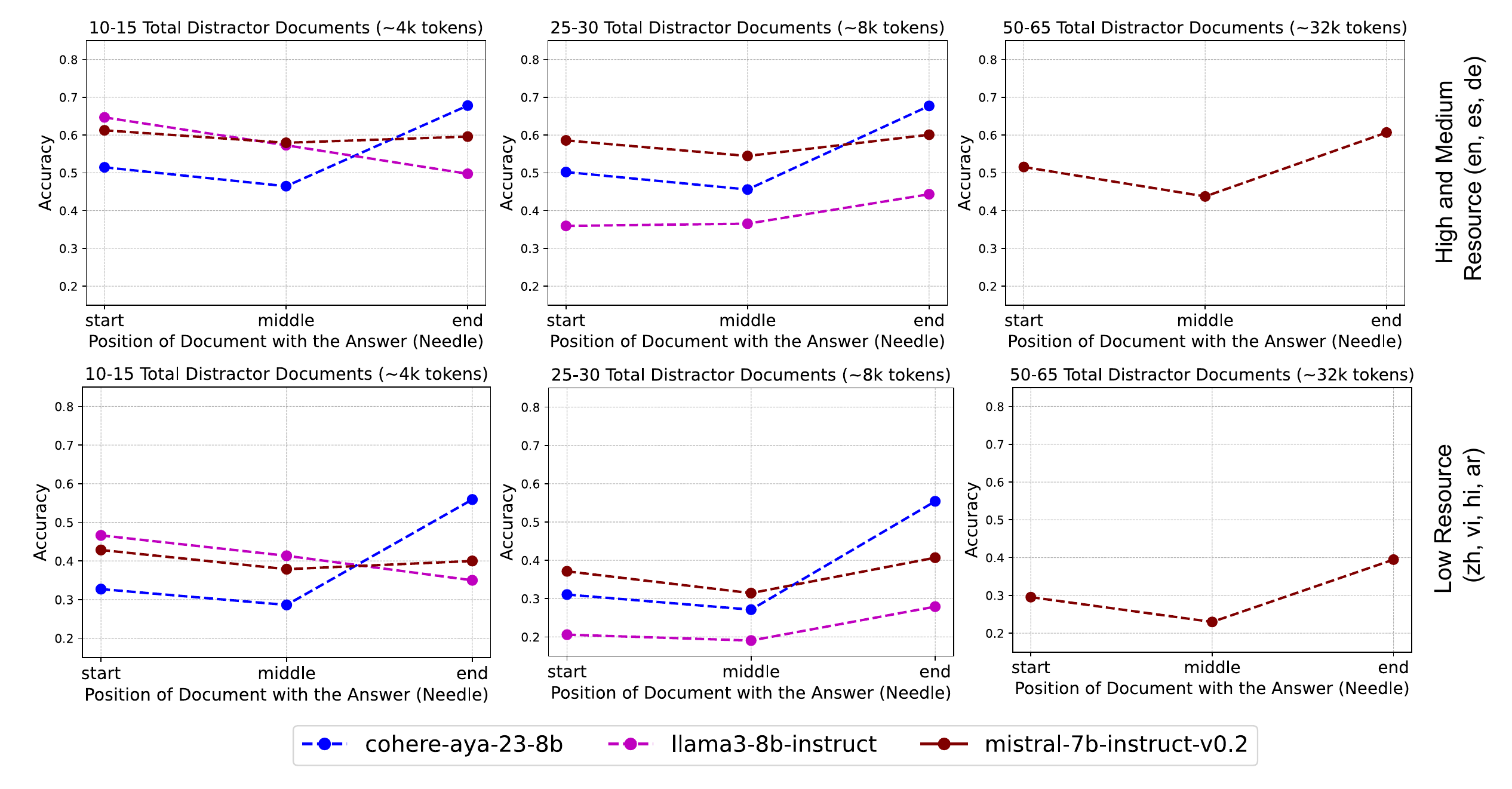}
\caption{Effect of changing position of answer document (needle).}
\label{fig:figure4}
\vspace{-0.5mm}
\end{figure*}

\begin{table}[t]
\small
\centering
\resizebox{0.9\columnwidth}{!}{%
\begin{tabular}{|c|ccccccc|}
\hline
\diagbox{$H$}{$N$} & en & de & es & zh & vi & hi & ar \\
\hline
\multicolumn{8}{c}{\cellcolor[HTML]{f3f3f3} Mistral-7B-Instruct-v0.2} \\
\hline
en & \cellcolor{red!30}\textcolor{black}{$0.68$} & \cellcolor{red!15}\textcolor{black}{$0.37$} & \cellcolor{red!15}\textcolor{black}{$0.35$} & \cellcolor{red!10}\textcolor{black}{$0.30$} & \cellcolor{red!10}\textcolor{black}{$0.31$} & \cellcolor{red!5}\textcolor{black}{$0.25$} & \cellcolor{red!5}\textcolor{black}{$0.24$} \\
de & \cellcolor{red!30}\textcolor{black}{$0.71$} & \cellcolor{red!15}\textcolor{black}{$0.37$} & \cellcolor{red!15}\textcolor{black}{$0.38$} & \cellcolor{red!5}\textcolor{black}{$0.24$} & \cellcolor{red!10}\textcolor{black}{$0.34$} & \cellcolor{red!5}\textcolor{black}{$0.24$} & \cellcolor{red!5}\textcolor{black}{$0.25$} \\
es & \cellcolor{red!30}\textcolor{black}{$0.70$} & \cellcolor{red!15}\textcolor{black}{$0.40$} & \cellcolor{red!15}\textcolor{black}{$0.39$} & \cellcolor{red!10}\textcolor{black}{$0.31$} & \cellcolor{red!10}\textcolor{black}{$0.34$} & \cellcolor{red!5}\textcolor{black}{$0.26$} & \cellcolor{red!5}\textcolor{black}{$0.28$} \\
zh & \cellcolor{red!30}\textcolor{black}{$0.73$} & \cellcolor{red!20}\textcolor{black}{$0.43$} & \cellcolor{red!15}\textcolor{black}{$0.39$} & \cellcolor{red!10}\textcolor{black}{$0.31$} & \cellcolor{red!10}\textcolor{black}{$0.35$} & \cellcolor{red!5}\textcolor{black}{$0.28$} & \cellcolor{red!5}\textcolor{black}{$0.27$} \\
vi & \cellcolor{red!30}\textcolor{black}{$0.75$} & \cellcolor{red!20}\textcolor{black}{$0.47$} & \cellcolor{red!15}\textcolor{black}{$0.42$} & \cellcolor{red!10}\textcolor{black}{$0.33$} & \cellcolor{red!10}\textcolor{black}{$0.37$} & \cellcolor{red!5}\textcolor{black}{$0.30$} & \cellcolor{red!5}\textcolor{black}{$0.27$} \\
hi & \cellcolor{red!40}\textcolor{black}{\textbf{$0.80$}} & \cellcolor{red!20}\textcolor{black}{$0.47$} & \cellcolor{red!20}\textcolor{black}{\textbf{$0.44$}} & \cellcolor{red!10}\textcolor{black}{\textbf{$0.36$}} & \cellcolor{red!10}\textcolor{black}{$0.38$} & \cellcolor{red!5}\textcolor{black}{$0.32$} & \cellcolor{red!10}\textcolor{black}{\textbf{$0.33$}} \\
ar & \cellcolor{red!30}\textcolor{black}{$0.76$} & \cellcolor{red!20}\textcolor{black}{\textbf{$0.49$}} & \cellcolor{red!15}\textcolor{black}{$0.43$} & \cellcolor{red!10}\textcolor{black}{$0.31$} & \cellcolor{red!15}\textcolor{black}{\textbf{$0.40$}} & \cellcolor{red!10}\textcolor{black}{\textbf{$0.33$}} & \cellcolor{red!5}\textcolor{black}{$0.30$} \\
\hline
\multicolumn{8}{c}{\cellcolor[HTML]{f3f3f3} Llama3-8B-Instruct} \\
\hline
en & \cellcolor{red!25}\textcolor{black}{$0.61$} & \cellcolor{red!10}\textcolor{black}{$0.21$} & \cellcolor{red!10}\textcolor{black}{$0.24$} & \cellcolor{red!10}\textcolor{black}{$0.21$} & \cellcolor{red!10}\textcolor{black}{$0.23$} & \cellcolor{red!10}\textcolor{black}{$0.23$} & \cellcolor{red!10}\textcolor{black}{$0.22$} \\
de & \cellcolor{red!25}\textcolor{black}{$0.64$} & \cellcolor{red!10}\textcolor{black}{$0.25$} & \cellcolor{red!10}\textcolor{black}{$0.25$} & \cellcolor{red!5}\textcolor{black}{$0.15$} & \cellcolor{red!10}\textcolor{black}{$0.20$} & \cellcolor{red!10}\textcolor{black}{$0.22$} & \cellcolor{red!5}\textcolor{black}{$0.19$} \\
es & \cellcolor{red!30}\textcolor{black}{$0.66$} & \cellcolor{red!15}\textcolor{black}{$0.28$} & \cellcolor{red!15}\textcolor{black}{$0.28$} & \cellcolor{red!10}\textcolor{black}{$0.19$} & \cellcolor{red!10}\textcolor{black}{$0.25$} & \cellcolor{red!10}\textcolor{black}{$0.24$} & \cellcolor{red!10}\textcolor{black}{$0.24$} \\
zh & \cellcolor{red!25}\textcolor{black}{$0.61$} & \cellcolor{red!10}\textcolor{black}{$0.22$} & \cellcolor{red!10}\textcolor{black}{$0.23$} & \cellcolor{red!10}\textcolor{black}{$0.19$} & \cellcolor{red!10}\textcolor{black}{$0.19$} & \cellcolor{red!10}\textcolor{black}{$0.20$} & \cellcolor{red!5}\textcolor{black}{$0.18$} \\
vi & \cellcolor{red!30}\textcolor{black}{$0.65$} & \cellcolor{red!15}\textcolor{black}{$0.29$} & \cellcolor{red!15}\textcolor{black}{$0.28$} & \cellcolor{red!10}\textcolor{black}{$0.18$} & \cellcolor{red!20}\textcolor{black}{\textbf{$0.30$}} & \cellcolor{red!10}\textcolor{black}{$0.24$} & \cellcolor{red!10}\textcolor{black}{$0.20$} \\
hi & \cellcolor{red!35}\textcolor{black}{\textbf{$0.70$}} & \cellcolor{red!20}\textcolor{black}{\textbf{$0.32$}} & \cellcolor{red!20}\textcolor{black}{\textbf{$0.32$}} & \cellcolor{red!15}\textcolor{black}{\textbf{$0.26$}} & \cellcolor{red!15}\textcolor{black}{$0.27$} & \cellcolor{red!20}\textcolor{black}{\textbf{$0.29$}} & \cellcolor{red!15}\textcolor{black}{\textbf{$0.26$}} \\
ar & \cellcolor{red!30}\textcolor{black}{$0.65$} & \cellcolor{red!15}\textcolor{black}{$0.27$} & \cellcolor{red!15}\textcolor{black}{$0.29$} & \cellcolor{red!10}\textcolor{black}{$0.22$} & \cellcolor{red!15}\textcolor{black}{$0.25$} & \cellcolor{red!15}\textcolor{black}{$0.25$} & \cellcolor{red!15}\textcolor{black}{$0.25$} \\
\hline
\multicolumn{8}{c}{\cellcolor[HTML]{f3f3f3} Cohere-Aya-23-8B} \\
\hline
en & \cellcolor{red!30}\textcolor{black}{$0.70$} & \cellcolor{red!20}\textcolor{black}{$0.51$} & \cellcolor{red!15}\textcolor{black}{$0.47$} & \cellcolor{red!10}\textcolor{black}{$0.36$} & \cellcolor{red!10}\textcolor{black}{$0.39$} & \cellcolor{red!5}\textcolor{black}{$0.33$} & \cellcolor{red!5}\textcolor{black}{$0.30$} \\
de & \cellcolor{red!30}\textcolor{black}{$0.71$} & \cellcolor{red!20}\textcolor{black}{\textbf{$0.53$}} & \cellcolor{red!15}\textcolor{black}{$0.46$} & \cellcolor{red!10}\textcolor{black}{$0.36$} & \cellcolor{red!15}\textcolor{black}{$0.41$} & \cellcolor{red!5}\textcolor{black}{$0.33$} & \cellcolor{red!10}\textcolor{black}{$0.33$} \\
es & \cellcolor{red!30}\textcolor{black}{$0.68$} & \cellcolor{red!20}\textcolor{black}{$0.48$} & \cellcolor{red!35}\textcolor{black}{\textbf{$0.62$}} & \cellcolor{red!10}\textcolor{black}{$0.35$} & \cellcolor{red!15}\textcolor{black}{$0.43$} & \cellcolor{red!10}\textcolor{black}{$0.35$} & \cellcolor{red!5}\textcolor{black}{$0.32$} \\
zh & \cellcolor{red!30}\textcolor{black}{$0.72$} & \cellcolor{red!20}\textcolor{black}{$0.47$} & \cellcolor{red!15}\textcolor{black}{$0.44$} & \cellcolor{red!35}\textcolor{black}{\textbf{$0.57$}} & \cellcolor{red!10}\textcolor{black}{$0.39$} & \cellcolor{red!20}\textcolor{black}{\textbf{$0.38$}} & \cellcolor{red!10}\textcolor{black}{$0.37$} \\
vi & \cellcolor{red!35}\textcolor{black}{\textbf{$0.73$}} & \cellcolor{red!20}\textcolor{black}{$0.46$} & \cellcolor{red!15}\textcolor{black}{$0.46$} & \cellcolor{red!10}\textcolor{black}{$0.31$} & \cellcolor{red!30}\textcolor{black}{\textbf{$0.49$}} & \cellcolor{red!10}\textcolor{black}{$0.36$} & \cellcolor{red!10}\textcolor{black}{$0.36$} \\
hi & \cellcolor{red!20}\textcolor{black}{$0.61$} & \cellcolor{red!15}\textcolor{black}{$0.42$} & \cellcolor{red!15}\textcolor{black}{$0.44$} & \cellcolor{red!10}\textcolor{black}{$0.34$} & \cellcolor{red!10}\textcolor{black}{$0.39$} & \cellcolor{red!10}\textcolor{black}{$0.35$} & \cellcolor{red!5}\textcolor{black}{$0.31$} \\
ar & \cellcolor{red!25}\textcolor{black}{$0.66$} & \cellcolor{red!15}\textcolor{black}{$0.45$} & \cellcolor{red!15}\textcolor{black}{$0.44$} & \cellcolor{red!10}\textcolor{black}{$0.35$} & \cellcolor{red!15}\textcolor{black}{$0.41$} & \cellcolor{red!10}\textcolor{black}{$0.37$} & \cellcolor{red!30}\textcolor{black}{\textbf{$0.50$}} \\
\hline
\end{tabular}%
}
\caption{Pairwise accuracy results of selected models (averaged across the context lengths) on the MLNeedle test. Language of relevant information and distractor documents is abbreviated as $N$ and $H$, respectively. We observe that performance is heavily influenced by the language of relevant information ($N$), whereas the language of distractor documents ($H$) plays a limited role. }
\label{tab:pairwise_accuracy}
\vspace{-5mm}
\end{table}

\section{Ablation Studies}
\label{sec:ablation}
\label{sec:ablation_analysis}

\textbf{Effect of Temperature Sampling.}
Here, we investigate whether the choice of generation strategy significantly influences the model's performance. Table \ref{tab:effect_of_temperature_sampling} compares the performance of Mistral-7B-Instruct-v0.2 under two different generation strategies: sampling with fixed parameters (temperature = $0.7$, top-$k$ = $50$) and greedy decoding. We observe that both strategies yield comparable results with minimal deviation across different context sizes. Furthermore, as shown in Figure \ref{fig:effect_of_temperature_sampling}, the overall accuracy tends to remain consistent regardless of the generation method employed. 

\textbf{Effect of Instruction Fine-tuning.}
To understand the impact of instruction fine-tuning on LLMs' use of multilingual long contexts, we compare the MLNeedle test performance of Mistral-7B-Instruct-v0.2 with its base variant (pre-instruction tuning) using the same experimental setup as in Section \ref{sec:MLNeedle_setup}. Table \ref{tab:effect_of_instruction_tuning} shows that Mistral-7B-Instruct-v0.2 consistently outperforms Mistral-7B-v0.1 across different context lengths. Instruction-tuning also reduces worst-case performance disparity from nearly $70\%$ to $30\%$. These findings align with prior work showing that instruction-tuning enhances cross-lingual knowledge alignment and improves information retrieval across languages \cite{shaham-multilingual-IT-2024, gao-etal-multilingual-IT-2024}.

\textbf{Effect of the Task Format.} The task of multilingual question answering through fact retrieval from a long context can be considered a challenging task for the choice of models under study. To prove the reliability of our results as being a property of long-context multilingual LMs, we also evaluate the models on a simpler secondary task — identifying the presence of relevant information within the input context. To this end, we defined the \textbf{existence accuracy} metric in Section \ref{sec: evaluation_metric}, which measures the proportion of samples where the model correctly identifies whether the relevant information is present in the provided passages.

Figure \ref{fig:ablation_mini} shows the results for Mistral-7B-Instruct-v0.2 across four different context lengths. We observe similar results for both Exact Accuracy and Existence Accuracy, showcasing that our findings are not solely because of the nature of the task. We also observe similar results when we vary the language of distractor passages in the input context. More details can be found in Appendix \ref{sec: appendix_ablation}.

\begin{figure}[t]
\centering \includegraphics[width=\columnwidth]{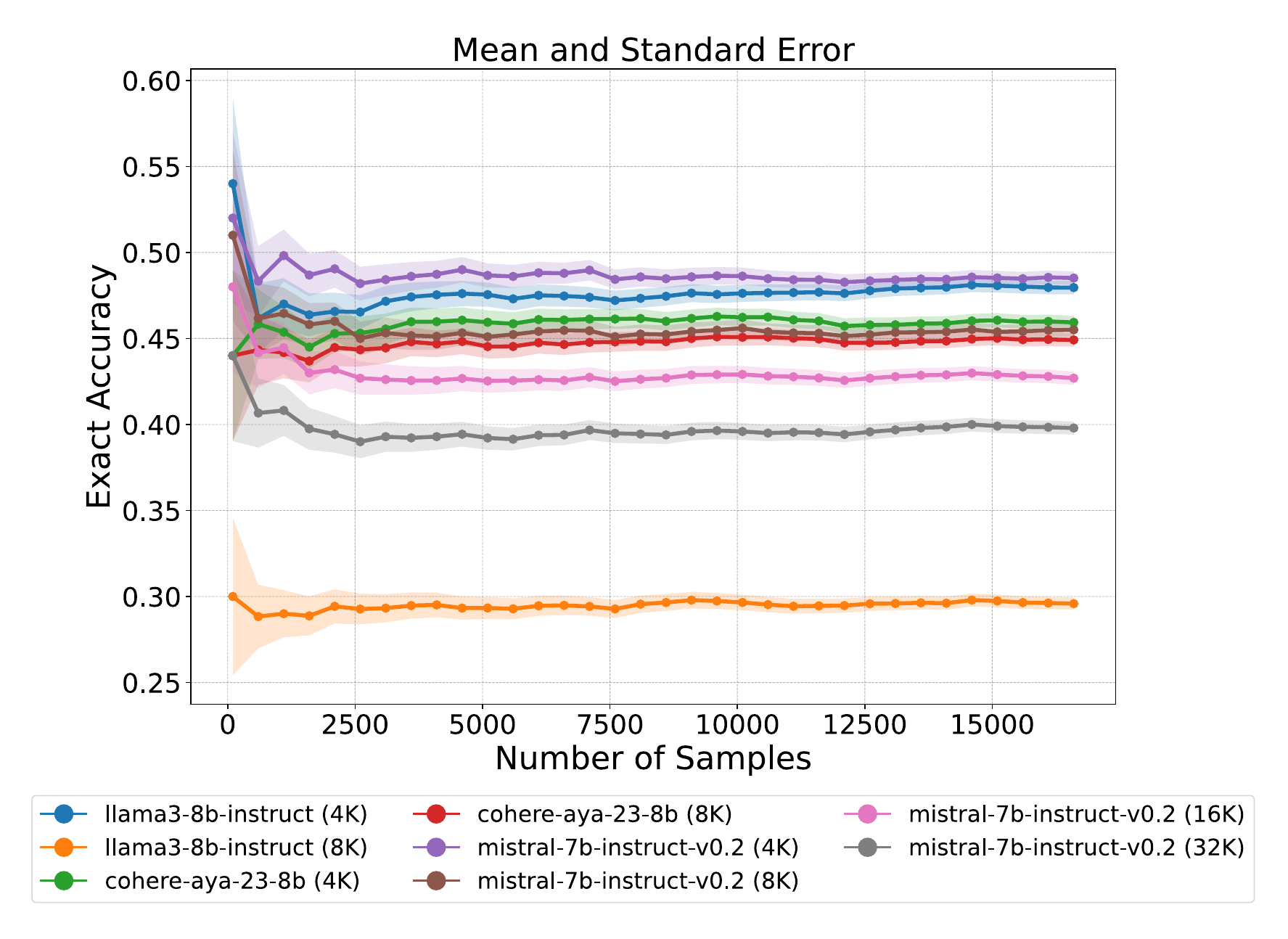}
\caption{Exact accuracy of models on varying sample sizes for evaluation. Solid lines denote the accuracy, and the shaded area denotes the standard error.}
\label{fig:stat_test}
\vspace{-3mm}
\end{figure}

\begin{figure}[t]
\includegraphics[width=\columnwidth]
{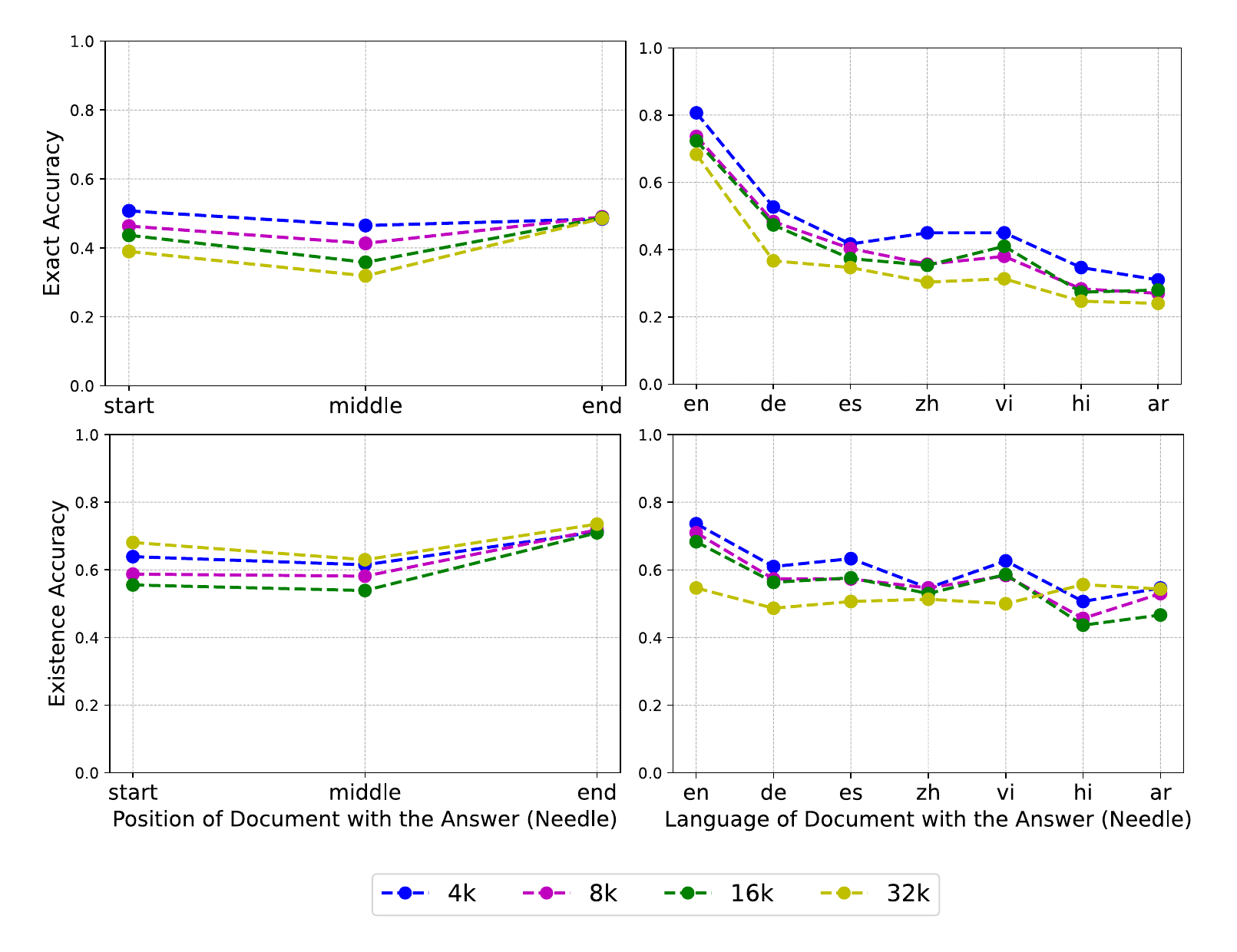}
\caption{Exact Accuracy and Existence Accuracy plots for Mistral-7B-Instruct-v0.2 across four different context lengths on varying (from left to left) : (i) Position of Document with the Answer (Needle) and (ii) Language of Document with the Answer (Needle).}
\label{fig:ablation_mini}
\vspace{-3.5mm}
\end{figure}

\textbf{Statistical Significance}: Figure \ref{fig:stat_test} presents the results of significance testing for selected models, computed over linearly increasing sample sizes from $100$ to $16,800$. We observe that accuracy stabilizes after approximately $2,500$ samples. Furthermore, a significant reduction in standard error is observed as the sample size increases from $100$ to $2,500$. These results underscore that a sample size of $2,500$ is sufficient to achieve reliable and consistent evaluation outcomes. Further details are provided in Appendix \ref{sec: appendix_ablation}.

\begin{table}[t]
\begin{center}
\resizebox{\columnwidth}{!}{%
\begin{tabular}{l|ccccccc}
\toprule
\textbf{Model} & \textbf{Baseline} & \textbf{4K} & \textbf{8K} & \textbf{16K} & \textbf{32K} & \textbf{Avg.} \\ 
\midrule
Mistral-Base & $0.383$ & $0.142$ & $0.100$ & $0.097$ & $0.102$ & $0.165$ \\
Mistral-Instruct & $0.586$ & $0.478$ & $0.453$ & $0.436$ & $0.398$ & $0.470$ \\
\midrule
\multicolumn{1}{c}{$\Delta_{\mathrm{model}^\dagger - \mathrm{model}^\ast}$} & \textcolor{ForestGreen}{$\uparrow0.203$} & \textcolor{ForestGreen}{$\uparrow0.336$} & \textcolor{ForestGreen}{$\uparrow0.353$} & \textcolor{ForestGreen}{$\uparrow0.339$} & \textcolor{ForestGreen}{$\uparrow0.296$} & \textcolor{ForestGreen}{$\uparrow0.305$} \\
\bottomrule
\end{tabular}%
}
\end{center}
\vspace{-2mm}
\caption{Effect of instruction fine-tuning.}
\label{tab:effect_of_instruction_tuning}
\vspace{-5mm}
\end{table}

\begin{table}[t]
\begin{center}
\resizebox{\columnwidth}{!}{%
\begin{tabular}{l|ccccccc}
\toprule
\textbf{Model} & \textbf{Baseline} & \textbf{4K} & \textbf{8K} & \textbf{16K} & \textbf{32K} & \textbf{Avg.} \\ 
\midrule
Mistral-Instruct$_{\mathrm{ (GD)}}$ & $0.586$ & $0.478$ & $0.453$ & $0.436$ & $0.398$ & $0.470$ \\
Mistral-Instruct$_{\mathrm{ (TS)}}$ & $0.580$ & $0.485$ & $0.455$ & $0.427$ & $0.398$ & $0.469$ \\
\midrule
\multicolumn{1}{c}{$\Delta_{\mathrm{model}^\dagger - \mathrm{model}^\ast}$} & \textcolor{red}{$\downarrow0.006$} & \textcolor{ForestGreen}{$\uparrow0.007$} & \textcolor{ForestGreen}{$\uparrow0.002$} & \textcolor{red}{$\downarrow0.009$} & \textcolor{blue}{$0.000$} & 
\textcolor{red}{$\downarrow0.001$} \\
\bottomrule
\end{tabular}%
}
\end{center}
\caption{Effect of temperature sampling vs greedy decoding.}
\label{tab:effect_of_temperature_sampling}
\vspace{-5mm}
\end{table}

\section{Related Work}

\paragraph{Multilingual Question Answering and Information Retrieval.}
Multilingual question answering (QA) and information retrieval (IR) have become increasingly important as LLMs are deployed in diverse linguistic environments \cite{shaham-multilingual-IT-2024}. Historically, QA datasets and benchmarks have been monolingual, primarily focusing on English. However, efforts such as MLQA \cite{lewis2020-MLQA} and XQuAD \cite{Artetxe-XQUAD-2019} have introduced datasets supporting multiple languages, facilitating cross-lingual evaluation of QA systems. Despite these advancements, much of the research in multilingual QA has focused on scenarios where the context is relatively short or where the question and context are in the same language \cite{Artetxe-XQUAD-2019,lewis2020-MLQA, MKQA-2021}. A significant gap persists between monolingual and multilingual QA performance, particularly when models encounter cross-lingual scenarios or low-resource languages \cite{towards-end-to-end-MLQA-2022, guo-cross-lingual-gap-2023}. The MLQA \cite{lewis2020-MLQA} and MKQA \cite{MKQA-2021} datasets, for instance, provide a foundation for evaluating cross-lingual extractive QA, but do not address the complexities introduced by long contexts. Our work extends this line of inquiry by examining how LLMs perform in retrieving relevant information from long multilingual contexts, where input context may span multiple documents — an increasingly common scenario in real-world applications.

\paragraph{Long-context Language Models.}
The ability of LLMs to effectively handle long input contexts is a critical area of research, underpinning tasks such as document summarization, long-form text generation, and multi-hop question answering \cite{qin2023-nlp_effectiveness_long-range, wang-Multimodal-NIAH-2024}. Transformer-based models \cite{vaswani-attention-2023}, traditionally limited by their quadratic complexity relative to sequence length, have spurred the development of various techniques to scale attention mechanisms and manage long contexts more efficiently \cite{dai-etal-2019-transformer, dao2022flashattentionfastmemoryefficientexact}. Recent innovations have extended context windows significantly, with models now capable of processing up to $100$K tokens in some cases. \citet{hsieh2024-Ruler} explored the practical usability of long-context LLMs and the effectiveness of different attention mechanisms. Despite the recent improvements, the practical utility of LLMs in long-context scenarios is often constrained by issues related to attention decay, memory management, and the ability to accurately retrieve relevant information from within extended sequences \cite{hsieh2024-Ruler, li2024-Longcontextllmsstrugglelong, qin2023-nlp_effectiveness_long-range}. \citet{Liu2023-LostITM} highlighted a significant challenge in this domain: a marked decline in LLM performance when relevant information is situated in the middle of a long context. This study revealed a ``U-shaped" performance curve, where models performed best when relevant information was at the beginning or end of the context, with performance dropping significantly for information located centrally. These findings underscore ongoing challenges in designing LLMs that can robustly handle long contexts, particularly in maintaining attention \cite{hsieh-found-in-the-middle-2024} and relevance \cite{hsieh2024-Ruler} across the entire sequence.

\paragraph{Long-context Benchmarks and Tasks.}
Benchmarking the ability of LLMs to handle long contexts is crucial for understanding their real-world applicability \cite{hsieh2024-Ruler, wang-Multimodal-NIAH-2024}. Previous benchmarks, such as ZeroSCROLLS \cite{shaham-etal-ZeroScrols-2023} LongBench \cite{bai-etal-Longbench-2024}, have provided insights into the limitations and potential of LLMs in processing extended input sequences. These benchmarks typically involve tasks such as long-document QA, multi-hop reasoning, and even multimodal retrieval \cite{Liu2023-LostITM, hsieh2024-Ruler, wang-Multimodal-NIAH-2024}. L-Eval \cite{An2023-LEvalIS} curates tests using realistic data, which is filtered manually to ensure quality. Infinite-Bench \citet{zhang-infinite-bench-2024} includes tasks with length greater than 100K tokens. However, these benchmarks primarily focus on monolingual English contexts, leaving a significant gap in understanding how LLMs perform in multilingual settings with long input sequences. For example, the MIRACL dataset \cite{zhang-Miracl-2022m} introduces a multilingual retrieval challenge but does not apply it to long-context scenarios.

Recent findings by \citet{zhao-largelanguagemodelshandle-2024} reveal that LLMs process multilingual content in three stages: first, they convert input into an English-centric representation, then process it during task solving, and finally, generate output in the original language. Our results extend this hypothesis by suggesting that models struggle with retrieving information when dealing with non-Latin languages (Figure \ref{fig:figure3}). 

Furthermore, a recent study \cite{hsieh-found-in-the-middle-2024} highlights the intrinsic positional attention bias in Transformer-based architectures \cite{vaswani-attention-2023}, where models disproportionately focus on tokens at the beginning and end of a sequence. This explains the U-shaped performance curve observed in Figure \ref{fig:figure4}. We further speculate that since the model's attention mechanisms are already biased due to positional encodings, the additional complexity of processing low-resource languages where the model may already have weaker representations could exacerbate the difficulty in retrieving and utilizing middle-positioned information. This might explain why the performance drop is more pronounced in non-Latin languages. Exploring this further is interesting future work.

\section{Conclusion}
In this study, we introduced the MultiLingual Needle in a Haystack (MLNeedle) test, designed to systematically evaluate the long-context capabilities of multilingual LLMs. Through a series of controlled experiments, we investigated how changes in the language and position of relevant information in a long context affect the LLMs' retrieval performance. Our findings revealed that LLMs exhibit significant sensitivity to both the language and position of the relevant information, particularly when the relevant content is in non-Latin languages or positioned in the middle of a long context. Conversely, the models demonstrated relative robustness to variations in the language of distractor passages, suggesting that the primary challenges lie in how LLMs process and retrieve the \textit{needle} from diverse linguistic contexts. These findings underscore the need for further research to enhance the multilingual capabilities of LLMs, particularly in handling long-context scenarios where relevant information may be dispersed across different languages and positions within the input. Our work represents a first step towards systematically evaluating the long-context behavior of multilingual LLMs.

\appendix
\newpage
\section{Process of retrieving the distractor documents}
\label{sec:noise_retrieval_process}
We randomly sample 10,000 Wikipedia passages from mMARCO for each of the languages under study. We encode every question in MLQA and every passage in mMARCO in a 768 dimensional dense embedding space using paraphrase-multilingual-mpnet-base-v2, trained on 50+ languages. Following that, we rank 300 most similar Wikipedia passages for every language against each question in MLQA using cosine similarity score.

\section{Prompt Templates}
\label{sec:prompt_templates}

In our experiments, we adopt a vanilla prompt template commonly used in multi-document question answering, following the linear \texttt{\textbf{<Instruction> + <Documents> + <Query>}} input sequence.
Table \ref{tab:prompt_template_exact_match_accuracy} and Table \ref{tab:prompt_template_existence_accuracy} highlight the respective prompt templates used for evaluating \texttt{exact\_accuracy} and \texttt{existence\_accuracy}. Additionally, Figure \ref{fig:figure5} and Figure \ref{fig:figure6} provide a example of actual input prompt for \texttt{exact\_accuracy} evaluation using three documents.

\section{Controlling Context Length}
\label{sec:controlling_context_length}
To control the size of the input context within provided context size proposed by the model specifications, we allow for maximal number of distractor passages to fit within the size allowance. Using mMARCO passages as distractor passages we observed following number of documents to appear in the input context for each context size : 4K - $\sim$10-15 distractor documents, 8K - $\sim$25-30 distractor documents, 32K - $\sim$50-65 distractor documents.

\section{Automated Evaluation Framework}
\label{appendix:evaluation_framework}
This section outlines the key steps involved in our automated evaluation process. To better understand our evaluation setup, we provide an example of a positive and negative instance from MLNeedle in Figure \ref{fig:evaluation_example}. 

For each input prompt, the model generates a prediction (\texttt{y\_pred}), which may be in a different language from the ground truth (\texttt{y\_true}). Each question-answer instance in MLNeedle dataset from MLQA has a golden answer provided in English, regardless of its language. To accurately compare \texttt{y\_pred} with \texttt{y\_true}, both must be in the same language. However, since \texttt{y\_pred} can be generated in any language, we translate all model predictions into English using Google Translate.

As shown in Figure \ref{fig:evaluation_example}, the translation step ensures that the prediction is in the same language as the ground truth, allowing for a direct and reliable comparison.
Once the predictions are translated, we proceed to compute the exact accuracy metric as defined in section \ref{sec: evaluation_metric}. This comparison helps us determine if the model has successfully identified and retrieved the correct information from the input context.

By translating all model outputs to English before comparison, we minimize the risk of false negatives due to language differences. This process ensures that all comparisons are reliable and that the evaluation accurately reflects the model's ability to retrieve relevant information, regardless of the input language.

\section{Additional Results}
\label{sec: appendix_ablation}
\subsection{Ablation Study: Effect of Evaluation Metric }\

\begin{figure}[ht]
\centering
\includegraphics[width=0.85\columnwidth]{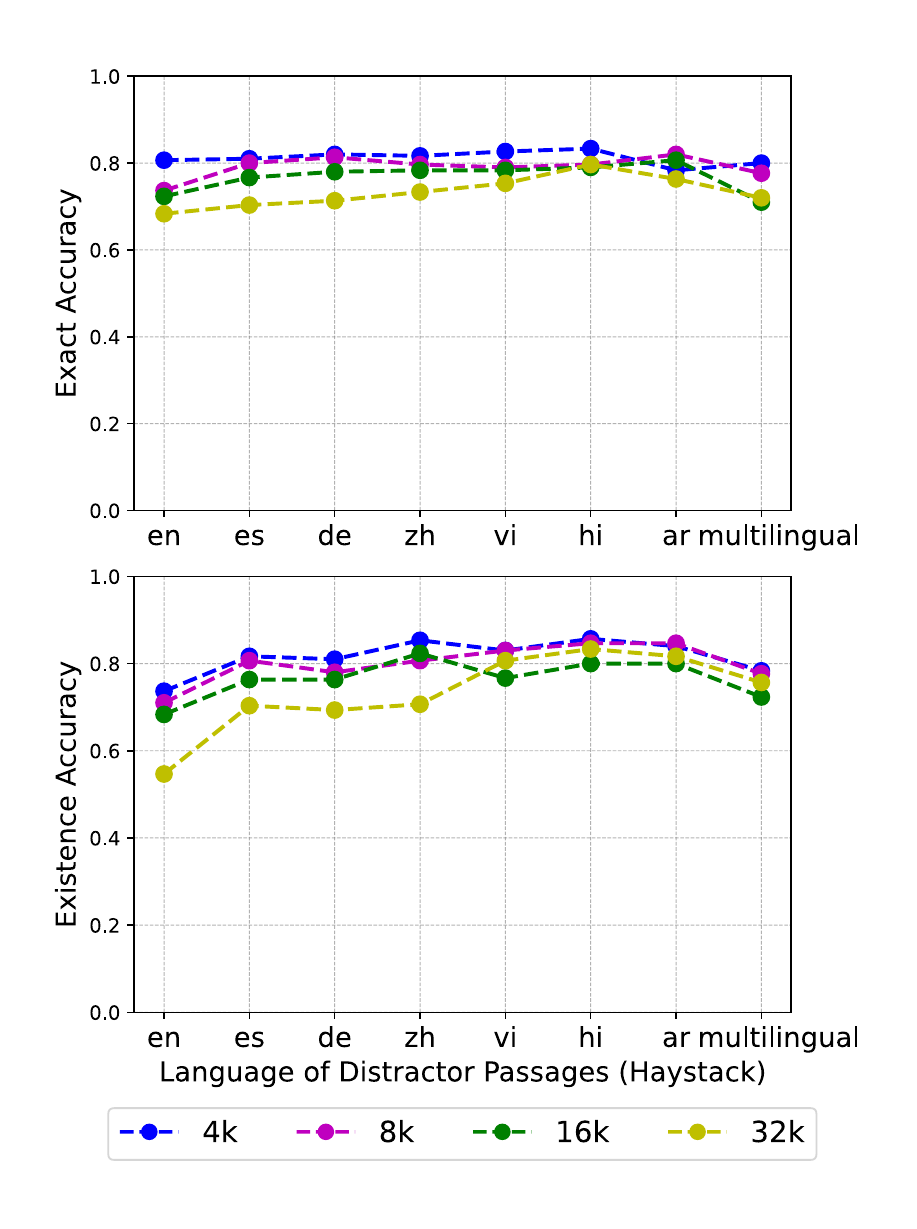}
\caption{Exact Accuracy and Existence Accuracy plots for Mistral-7B-Instruct-v0.2 across four different context lengths on the varying language of distractor passages (Haystack).}

\label{fig:ablation_mini_appendix}
\vspace{-0.25mm}
\end{figure}

In Section \ref{sec:ablation}, we discussed how we obtained similar plots for `Exact Accuracy' and `Existence Accuracy'. From Figure \ref{fig:ablation_mini_appendix}, we observe the similarity between plots obtained on changing the language of the distractor documents in the input context. Both `Existence Accuracy' and `Exact Accuracy' do not vary in changing the language of the distractor documents in the input context.

\begin{figure}[t]
\includegraphics[width=\columnwidth]{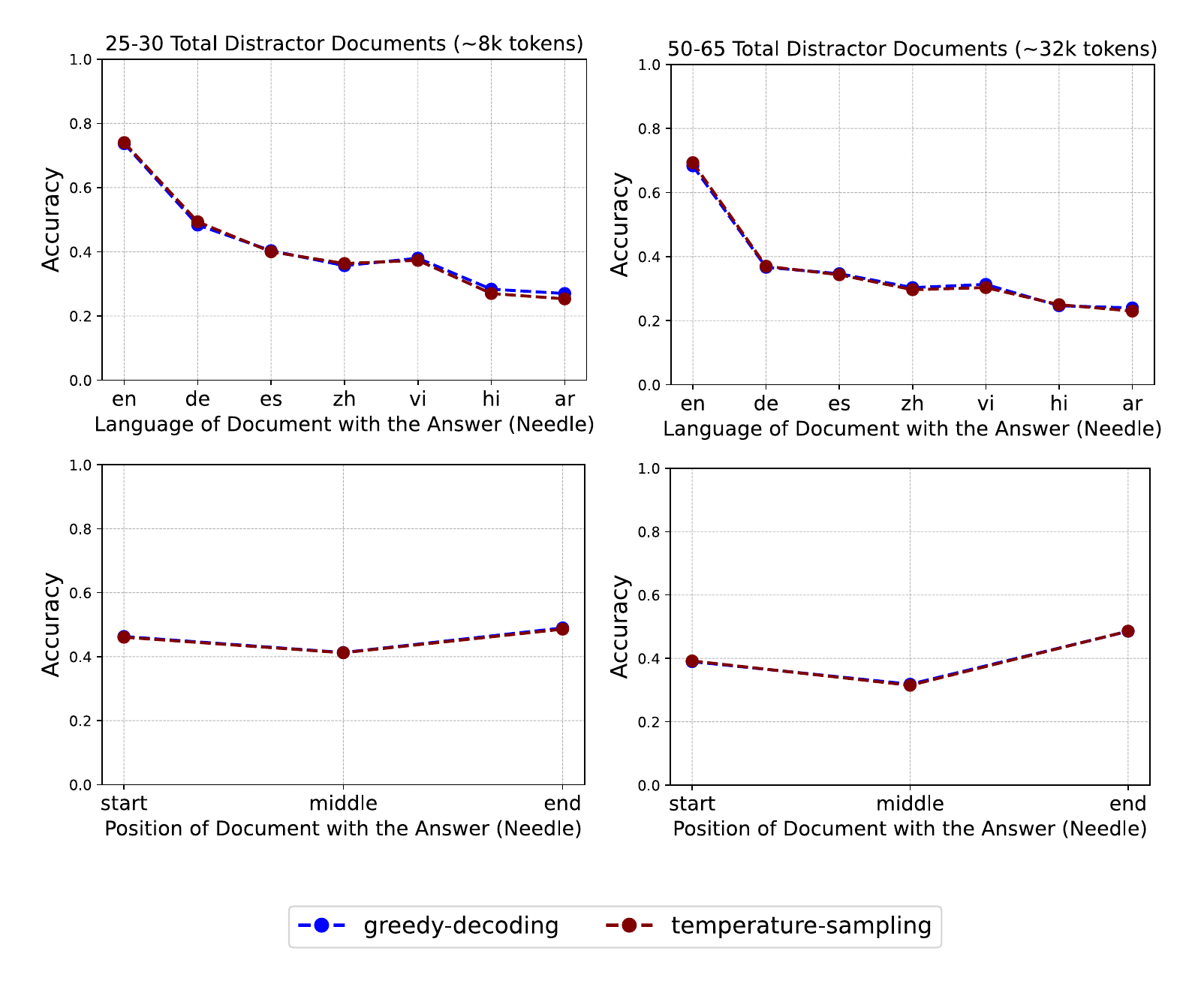}
\caption{Performance of mistral-7b-instruct-v0.2 using different decoding strategies. We report similar trends for two different context lengths : $\sim$8K tokens and  $\sim$32K tokens.}
\label{fig:effect_of_temperature_sampling}
\vspace{-5mm}
\end{figure}

\subsection{Statistical Significance}
To ensure the reliablity of our evaluation, we conduct a hypothesis test for exact accuracy as defined in section \ref{sec: evaluation_metric}. We conduct the test for selected models under a binomial distribution \(\text{Binomial}(1, p)\), where \(p\) is the probability of success on an individual trial. The standard error ($SE$) of this test is computed as follows:

\begin{equation}
\begin{aligned}
SE = \sqrt{\frac{p(1 - p)}{s}},
\end{aligned}
\end{equation}

where \(s\) is the number of trials (evaluation samples). We vary the sample size linearly, starting from $100$ samples and increasing by $500$ samples at each step, i.e., $100$, $600$, $1100$, $1600$, and so on, until reaching the full sample size of MLNeedle ($16,800$ samples). Note that at each step, we randomly select the samples from the MLNeedle dataset. Figure \ref{fig:stat_test} highlights the results of our statistical tests. We observe that the exact accuracy stabilizes after approximately $2,500$ samples, and the standard error decreases significantly as the sample size increases from $100$ to $16,800$. 

\begin{figure*}[!ht]
\includegraphics[width=\textwidth,keepaspectratio,trim={0 0 0 0},clip]{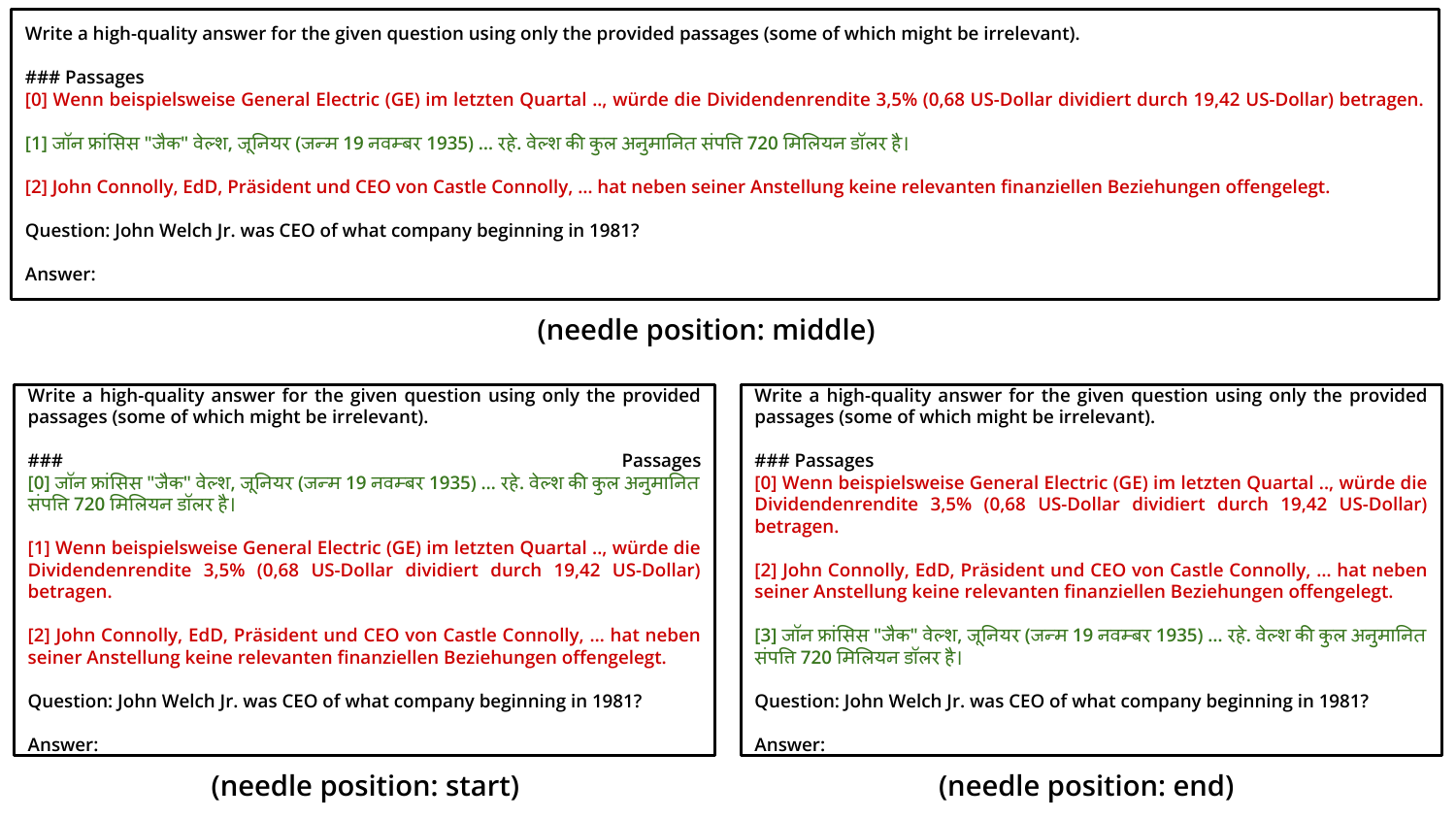}
\caption{Modulating the position of relevant information (`needle') within the input context (`haystack') presented in Figure \ref{fig:figure5}.}
\label{fig:figure6}
\vspace{-0.25mm}
\end{figure*}

\begin{table*}[!ht]
\begin{center}
\small 
\setlength{\tabcolsep}{2pt} 
\renewcommand{\arraystretch}{1.0} 
\resizebox{\textwidth}{!}{%
\begin{tabular}{p{1.3in}p{5.2in}}
\toprule
\textbf{\texttt{Prompt Template for Exact Match}} &
  \begin{tabular}[c]{@{}p{5.2in}@{}}
  \texttt{Write a high-quality answer for the given question using only the provided passages (some of which might be irrelevant).}\\\\
  \texttt{\#\#\# Passages}\\
  \texttt{\{input\_passages\}}\\\\
  \texttt{Question: \{question\}}\\\\
  \texttt{Answer:}
  \end{tabular} \\
\bottomrule
\end{tabular}%
}
\end{center}
\caption{Prompt template used for evaluating \texttt{exact\_accuracy}. The model is asked to generate an answer based solely on the provided passages.}
\label{tab:prompt_template_exact_match_accuracy}
\end{table*}

\begin{table*}[!ht]
\begin{center}
\small 
\setlength{\tabcolsep}{2pt} 
\renewcommand{\arraystretch}{1.0} 
\resizebox{\textwidth}{!}{%
\begin{tabular}{p{1.3in}p{5.2in}}
\toprule
\textbf{\texttt{Prompt Template for Existence Match}} &
  \begin{tabular}[c]{@{}p{5.2in}@{}}
  \texttt{Read the following list of passages and indicate whether any of the passages contain the right answer for the given question. Format your output strictly as 'Yes' or 'No'.}\\\\
  \texttt{\#\#\# Passages}\\
  \texttt{\{input\_passages\}}\\\\
  \texttt{Question: \{question\}}\\\\
  \texttt{Answer [Yes/No]:}
  \end{tabular} \\
\bottomrule
\end{tabular}%
}
\end{center}
\caption{Prompt template used for evaluating \texttt{existence\_accuracy}. The model is asked to determine if the correct answer is present within the provided passages.}
\label{tab:prompt_template_existence_accuracy}
\end{table*}

\begin{figure*}[!ht]
\includegraphics[width=\textwidth,keepaspectratio,trim={0 0 0 0},clip]{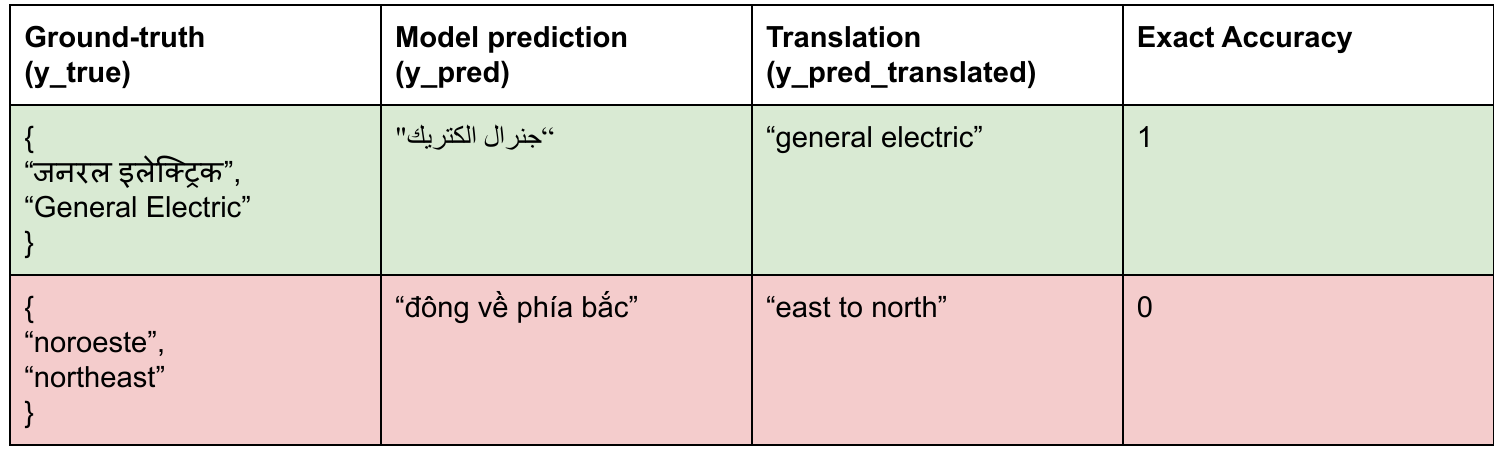}
\caption{An example of a positive and a negative instance while computing exact accuracy. The translation step enables a direct comparison between y\_true and y\_pred, which in turn helps reduce false-negatives.}
\label{fig:evaluation_example}
\vspace{-5mm}
\end{figure*}

\end{document}